\documentclass[10pt,twocolumn,letterpaper]{article}

\usepackage{iccv}              
\usepackage[accsupp]{axessibility}  

\usepackage[utf8]{inputenc} %
\usepackage[T1]{fontenc}    %
\usepackage{url}            %
\usepackage{booktabs}       %
\usepackage{amsfonts}       %
\usepackage{nicefrac}       %
\usepackage{microtype}      %
\usepackage{xcolor}         %

\usepackage{bbding}
\usepackage{multirow}
\usepackage{algorithm}
\usepackage[noend]{algorithmic}
\usepackage{graphicx, amsmath, amssymb, caption, subcaption, overpic, textpos}
\usepackage{arydshln} %
\usepackage{listings}
\usepackage{setspace}

\usepackage{xspace}

\newcommand{\tabref}[1]{Tab.~\ref{#1}}

\def\eg{\emph{e.g}\onedot} 
\def\ie{\emph{i.e}\onedot} 
 
 \def\vs{\emph{vs}\onedot}

\newcommand{\figref}[1]{Fig.~\ref{#1}}
\newcommand{\secref}[1]{Sec.~\ref{#1}}

\newlength\savewidth\newcommand\shline{\noalign{\global\savewidth\arrayrulewidth
  \global\arrayrulewidth 1pt}\hline\noalign{\global\arrayrulewidth\savewidth}}
\newcommand{\tablestyle}[2]{\setlength{\tabcolsep}{#1}\renewcommand{\arraystretch}{#2}\centering\footnotesize}

\usepackage{pifont}
\definecolor{mygreen}{RGB}{0,200,0}
\newcommand{\cmark}{\textcolor{mygreen}{\ding{51}}}%
\newcommand{\xmark}{\textcolor{red}{\ding{55}}}%
\usepackage{colortbl}
\definecolor{finalcolor}{gray}{.9}
\newcommand{\final}[1]{\cellcolor{finalcolor}{#1}}

\usepackage{subcaption}

\newcommand{\tokenizername}{TA-TiTok\xspace}
\newcommand{\modelname}{FlowTok\xspace}

\definecolor{iccvblue}{rgb}{0.21,0.49,0.74}
\usepackage[pagebackref,breaklinks,colorlinks,allcolors=iccvblue]{hyperref}

\def\paperID{3306} 
\def\confName{ICCV}
\def\confYear{2025}

\title{\modelname: Flowing Seamlessly Across Text and  Image Tokens}

\author{
  Ju He\textsuperscript{1} \quad  Qihang Yu\textsuperscript{1} \quad
  Qihao Liu\textsuperscript{2} \quad Liang-Chieh Chen\textsuperscript{1}
  \\
  \textsuperscript{1} ByteDance Seed \qquad
   \textsuperscript{2} Johns Hopkins University
    \\
   \url{https://tacju.github.io/projects/flowtok.html}
}
\begin{document}

\twocolumn[
{%
\maketitle\centering
\vspace{-26pt}
\includegraphics[width=0.97\linewidth]{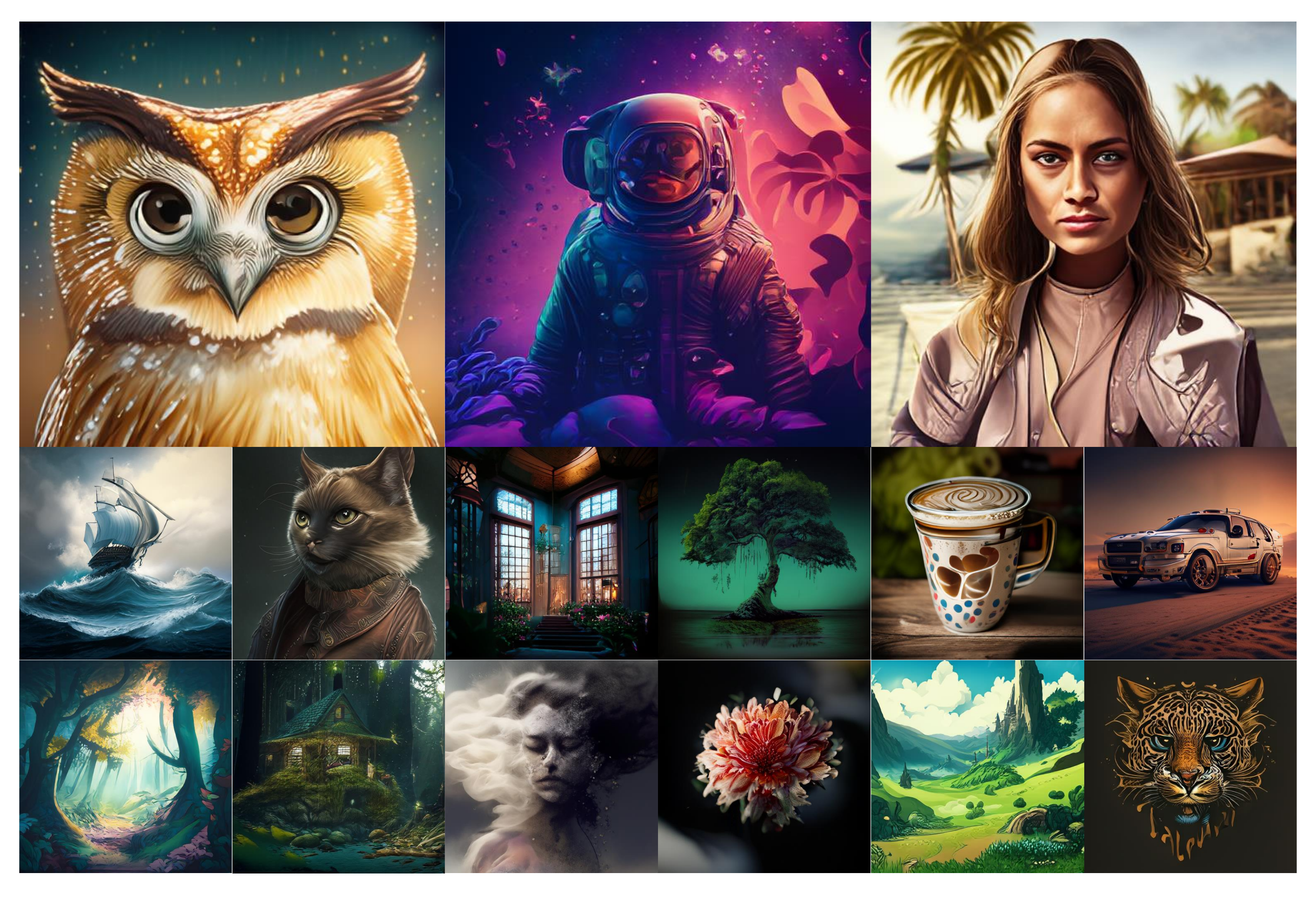}
  \vspace{-12pt}
  \captionof{figure}{
  \textbf{Text-to-Image Generation Results by \modelname.} \modelname projects both text and images into a unified, compact 1D latent space, enabling direct flow matching between 1D tokens and facilitating the efficient generation of diverse, high-fidelity images.
  }
  \vspace{10pt}
  \label{fig:teaser}
}
]

\begin{abstract}
Bridging different modalities lies at the heart of cross-modality generation. While conventional approaches treat the text modality as a conditioning signal that gradually guides the denoising process from Gaussian noise to the target image modality, we explore a much simpler paradigm—directly evolving between text and image modalities through flow matching. This requires projecting both modalities into a shared latent space, which poses a significant challenge due to their inherently different representations: text is highly semantic and encoded as 1D tokens, whereas images are spatially redundant and represented as 2D latent embeddings.
To address this, we introduce \modelname, a minimal framework that seamlessly flows across text and images by encoding images into a compact 1D token representation. Compared to prior methods, this design reduces the latent space size by 3.3$\times$ at an image resolution of 256, eliminating the need for complex conditioning mechanisms or noise scheduling. Moreover, \modelname naturally extends to image-to-text generation under the same formulation.
With its streamlined architecture centered around compact 1D tokens, \modelname is highly memory-efficient, requires significantly fewer training resources, and achieves much faster sampling speeds—all while delivering performance comparable to state-of-the-art models.
Code is available at \href{https://github.com/TACJu/FlowTok}{https://github.com/TACJu/FlowTok}.
\end{abstract}

\vspace{-40pt}
\begin{figure*}[t!]
  \centering
   \includegraphics[width=\linewidth]{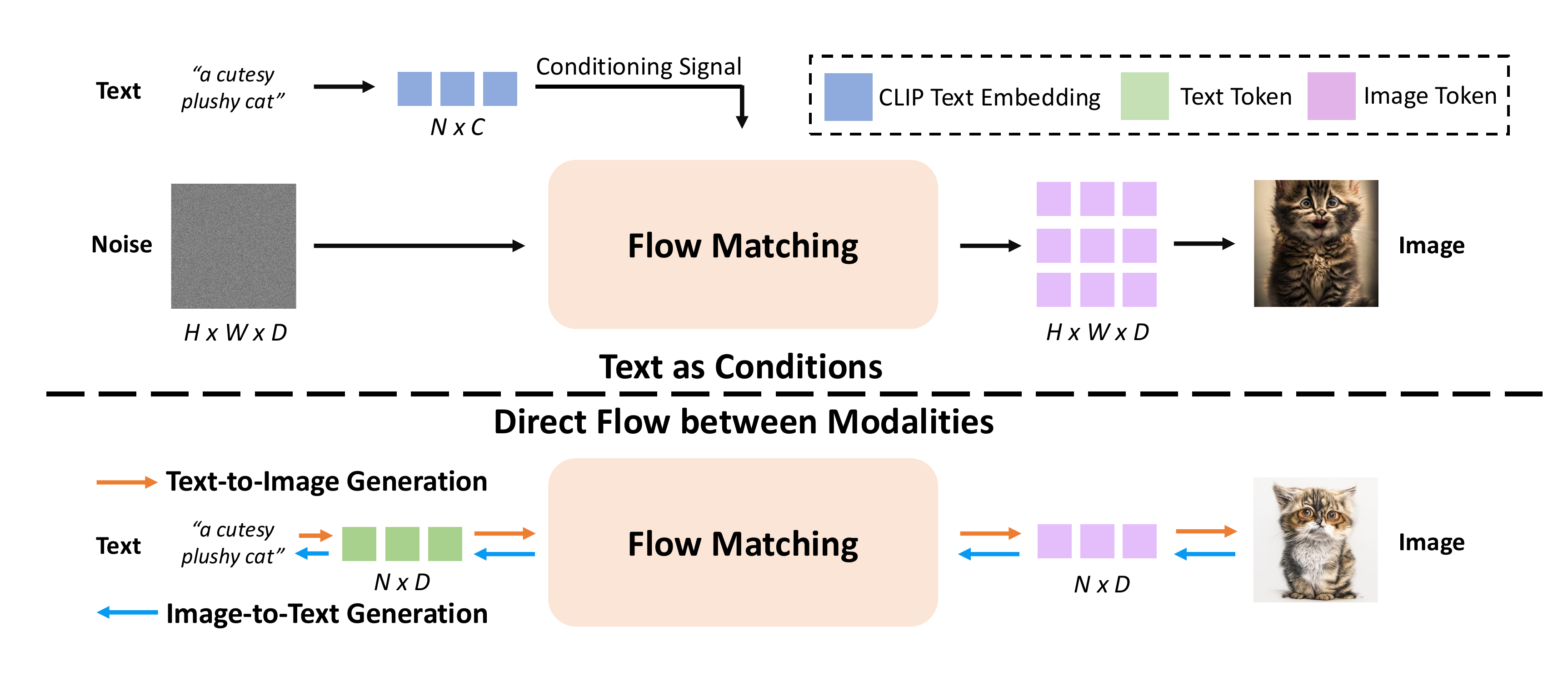}
   \caption{
   \textbf{Text as Conditions \vs Direct Flow between Modalities.}
   \textit{Top:} Conventional text-to-image generation relies on the diffusion process, where text serves as a conditioning signal to guide the denoising process.
   \textit{Bottom:} The proposed \modelname enables direct flow between text and image modalities by projecting both into a shared, compact 1D latent space, facilitating seamless generation of both.
   }
   \label{fig:vis_teaser}
\end{figure*}

\section{Introduction}
\label{sec:intro}

Bridging different modalities is essential for comprehending the diverse forms of data that represent our world, encompassing both understanding and generation.
In multimodal understanding, extensive research has focused on designing architectures that project different modalities into a shared latent space~\cite{li2019visualbert,lu2019vilbert,radford2021learning,jia2021scaling,cho2021unifying,wang2021simvlm,li2022blip,jaegle2021perceiver,alayrac2022flamingo,driess2023palm,he2024a}. These approaches have significantly advanced cross-modal representation learning and real-world understanding by leveraging a common latent space between modalities.

In contrast, multimodal generation (\eg, text-to-image generation) follows a different paradigm, primarily relying on the diffusion process~\cite{sohl2015deep,song2019generative,ho2020denoising,peebles2023scalable,liu2024alleviating}, where the source modality (\eg, text) serves as a conditioning signal to guide the denoising process.
Various conditioning mechanisms have been explored, including concatenation~\cite{bao2023all}, cross-attention~\cite{rombach2022high,chen2024pixartalpha}, conditioning embeddings~\cite{peebles2023scalable}, and hybrid strategies~\cite{esser2024scaling,kim2025democratizing}. While effective, these approaches introduce substantial complexity, requiring intricate conditioning mechanisms and noise scheduling.
This naturally raises an important question:
\textit{Can we unify multimodal understanding and generation by enabling direct transitions within a shared latent space?}

To address this question, we revisit flow matching~\cite{lipmanflow,albergo2022building,liuflow}—a modern generative framework that learns a direct path from noise to data, enabling faster convergence and accelerated sampling, leading to state-of-the-art multimodal generation results~\cite{esser2024scaling,flux2024}.
Unlike diffusion models, flow matching is not constrained to using noise as the source distribution; instead, it only requires the source and target distributions to share the same shape. Pioneering works~\cite{albergo2022building,liu2023i2sbimagetoimageschrodingerbridge,shi2023diffusion,tong2023simulation,zhou2023denoising} have demonstrated its effectiveness in learning direct mappings within the same modality (\eg, image-to-image generation).
Meanwhile, CrossFlow~\cite{liu2024flowing} extends flow matching to cross-modal learning by mapping text into a 2D latent space to match the shape of image embeddings, paving the way for new possibilities.
However, while this approach simplifies the overall pipeline, it still operates on 2D latent representations.
As a result, the additional computational overhead introduced by the text variational autoencoder~\cite{kingma2013auto} in CrossFlow makes it slower than modern text-to-image diffusion models like SD1.5 and SD2.1~\cite{rombach2022high}, ultimately contradicting its original goal of efficiency.

To this end, we introduce \modelname, a minimal framework that enables seamless \textbf{Flow}ing of \textbf{Tok}ens across text and image—the two most prevalent modalities (\cref{fig:vis_teaser}).
At the core of \modelname, both text and images are encoded into compact 1D latent tokens within a unified space, enabling direct flow matching between them. On the text side, \modelname employs a pre-trained text encoder~\cite{radford2021learning} to extract initial 1D text embeddings. Since these embeddings typically reside in a higher-dimensional space than image latents, \modelname introduces a lightweight text projector to map text embeddings into a low-dimensional variational latent space.
On the image side, \modelname builds on recent advancements in image tokenization~\cite{yu2024image,kim2025democratizing} to encode images into compact 1D latent tokens. Specifically, we enhance \tokenizername~\cite{kim2025democratizing} by integrating RoPE~\cite{su2021roformer} and SwiGLU FFN~\cite{shazeer2020glu}, improving positional information handling and reconstruction quality. To enable direct flow matching, we align the number of image latent tokens $K$ in \tokenizername to match the text encoder’s output sequence length ($K=77$ for CLIP text encoder).


By integrating these simple yet effective designs across text and image modalities, \modelname represents both in the same 1D low-dimensional space with shape $77 \times 16$ (77 tokens, each with 16 dimensions). This compact representation is 3.3$\times$ smaller than typical 2D flow matching shapes~\cite{sdvae} of $32 \times 32 \times 4$ for image resolutions of 256. This alignment enables fast, direct flow matching and seamless evolution between the two modalities.

Unlike standard flow matching models~\cite{ma2024sit,lipmanflow,liuflow}, \modelname eliminates the need for intricate conditioning mechanisms, offering a fully self-attention-based generative model. This allows for direct flow across modalities without additional complexity. Unlike CrossFlow~\cite{liu2024flowing}, which converts text into 2D embeddings, \modelname retains the 1D structure of text embeddings, avoiding the need for flattening and transformation into 2D. This simplifies the framework while eliminating reliance on heavy parametric contrastive losses for semantic preservation.



\begin{figure}[t!]
    \centering
    \begin{subfigure}[t]{0.45\linewidth}
        \centering
        \includegraphics[width=\linewidth]{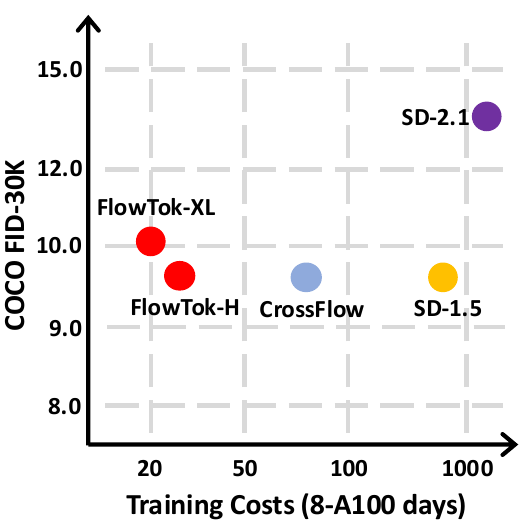}
        \caption{\textbf{FID \vs Training Costs.}}
        \label{fig:fidvst}
    \end{subfigure}%
    ~ 
    \begin{subfigure}[t]{0.45\linewidth}
        \centering
        \includegraphics[width=\linewidth]{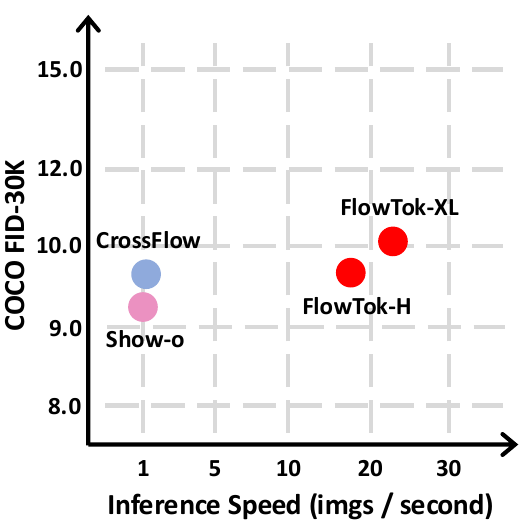}
        \caption{\textbf{FID \vs Inference Speed.}}
        \label{fig:fidvsi}
    \end{subfigure}
    \caption{
    \textbf{COCO Results.}
    \modelname presents comparable performance to previous methods on COCO while significantly reducing training resource requirements (\figref{fig:fidvst}) and achieving much faster sampling speed (\figref{fig:fidvsi}). This efficiency stems from its minimalist design centered around 1D tokens, which facilitates direct transformation between text and image modalities, leading to superior performance with enhanced computational efficiency.
    We note that the compared CrossFlow~\cite{liu2024flowing} uses high-quality proprietary data.
    }
    \label{fig:fid}
\end{figure}

As a result, \modelname offers a streamlined and resource-efficient training process. Its largest variant, \modelname-H (1.1B), supports a batch size of 8K on 8 A100 GPUs without requiring gradient checkpointing or gradient accumulation. In contrast, recent text-to-image models of similar scale typically require 32 to 64 A100 GPUs to train with a batch size of only 2K~\cite{rombach2022high,chen2024pixartalpha}. Moreover, \modelname converges significantly faster, as shown in~\figref{fig:fidvst}, with \modelname-H completing training in just 26.1 8-A100 days—far less than SD 2.1~\cite{rombach2022high}, which requires 1041.6 8-A100 GPU days.


Inference is also highly efficient, with \modelname achieving over 10$\times$ the throughput of Show-o~\cite{xie2024show} and CrossFlow~\cite{liu2024flowing}, as shown in~\figref{fig:fidvsi}. This dramatically reduces computational costs, making text-to-image research far more accessible. To ensure full reproducibility, we train \modelname exclusively on publicly available datasets, avoiding reliance on high-quality proprietary data. Remarkably, despite its minimalist design and reduced data requirements, \modelname achieves state-of-the-art text-to-image performance.


Beyond text-to-image generation, \modelname seamlessly extends to image-to-text generation, maintaining strong performance under the same minimalist framework. We believe our work establishes a strong foundation for future research in generalized cross-modality generation.


\section{Related Work}
\label{sec:related}

\noindent \textbf{Flow Matching.} 
Flow matching~\cite{lipmanflow,albergo2022building,liuflow} models generative processes by constructing a transport map between two distributions via an ordinary differential equation (ODE). It has recently gained traction as the foundation for state-of-the-art text-to-image and text-to-video synthesis models~\cite{esser2024scaling,polyak2024moviegencastmedia,chen2025goku,ren2024flowar,ren2025grouping}, offering faster training and sampling compared to conventional diffusion methods~\cite{ho2020denoising,peebles2023scalable,liu2024alleviating,liu2025revision}. Several works further optimize flow trajectories by minimizing curvature~\cite{lee2023minimizing,tong2023improving,pooladian2023multisample}.
Despite its theoretical flexibility in handling arbitrary distributions, recent approaches primarily evolve noise into target distributions, often relying on complex control signal conditioning, which complicates the pipeline and overlooks the potential of directly transforming control signals into target distributions. 
In contrast, only a few works~\cite{liuflow,albergo2022building,liu2023i2sbimagetoimageschrodingerbridge,shi2023diffusion,tong2023simulation,zhou2023denoising} explore direct transport within the same modality (\eg, image-to-image~\cite{liuflow,zhou2023denoising,fischer2023boosting}), leaving cross-modal transport (\eg, text-to-image) underexplored.
In this work, we introduce \modelname, a minimal yet effective framework that enables seamless flow across text and image modalities using 1D tokens. Unlike CrossFlow~\cite{liu2024flowing}, which follows a similar paradigm but relies on 2D latent representations and incurs additional computational costs due to the text variational encoder, \modelname operates within a unified, compact 1D token space. This design achieves a 3.3$\times$ compression rate in latent size, significantly reducing training costs and accelerating the sampling process, all while maintaining state-of-the-art performance.

\noindent \textbf{Text-to-Image Generation.}
Text-to-image generation has advanced rapidly in recent years, driven by various generative paradigms, including diffusion models~\cite{rombach2022high,saharia2022photorealistic,podell2023sdxl,chen2024pixartalpha}, flow matching models~\cite{esser2024scaling,chen2025goku,yu2025representation,yang20241,shin2025deeply}, sequence models~\cite{yu2022scaling,gafni2022make,fan2024fluid,yu2024randomized,ren2025beyond}, and masked generative models~\cite{chang2023muse,weber2024maskbit,bai2025meissonic,kim2025democratizing}.
While early works in each category establish the foundation for their respective approaches, subsequent advancements across different model types have primarily emerged from three key areas: careful data collection and advanced image recaptioning for improved data quality~\cite{betker2023improving,kim2025democratizing,deng2025coconut}, architectural and conditioning improvements for faster convergence and better text-image alignment~\cite{chen2024pixartalpha,esser2024scaling,fan2024fluid,chen2025goku}, and micro-conditioning for finer control over generated samples~\cite{podell2023sdxl,bai2025meissonic,kim2025democratizing}.
By contrast, this work introduces a minimalist framework \modelname that directly maps text tokens to image tokens, eliminating the need for noise scheduling and complex conditioning mechanisms. This streamlined design enhances both efficiency and simplicity while maintaining competitive performance.
\section{Preliminary}
\label{sec:preliminary}
\noindent \textbf{1D Visual Tokenization~\cite{yu2024image,kim2025democratizing}} deviates from traditional 2D grid-based latent tokenization by adopting a compact 1D representation, eliminating the need to preserve the 2D spatial structure. This work focuses on continuous 1D visual tokens for flow matching.
During tokenization, given an input image \(\mathbf{I} \in \mathbb{R}^{H \times W \times 3}\), the image is downscaled by a factor of $f$, resulting in patches $\mathbf{P} \in \mathbb{R}^{\frac{H}{f} \times \frac{W}{f} \times D}$.
These patches are concatenated with a set of latent tokens $\mathbf{L} \in \mathbb{R}^{K \times D}$to form a sequence that is passed through a Vision Transformer (ViT)~\cite{dosovitskiy2020image} encoder, $Enc$, to generate embeddings. Only the embeddings corresponding to the latent tokens are retained, forming a compact 1D latent representation. This representation is modeled as a Gaussian distribution with KL divergence regularization, resulting in a compact 1D VAE representation, $\mathbf{Z}_{I} \in \mathbb{R}^{K \times D}$.
In the de-tokenization phase, text guidance is applied by incorporating text embeddings generated by a pre-trained text encoder~\cite{radford2021learning}. These text embeddings are projected through a linear layer to align with the channel dimensions of ViT decoder, resulting in \(\mathbf{T} \in \mathbb{R}^{N \times D}\), where $N$ is the number of context tokens predefined by the text encoder. The text embedding $\mathbf{T}$ is then concatenated with the latent tokens $\mathbf{Z}_{\text{I}}$ and a set of mask tokens $\mathbf{M} \in \mathbb{R}^{\frac{H}{f} \times \frac{W}{f} \times D}$. The combined sequence is passed through the decoder $Dec$, yielding the reconstructed image \(\mathbf{\hat{I}}\).
Formally, with $\oplus$ denoting concatenation, the tokenization and de-tokenization can be represented as:
\begin{gather*}
    \mathbf{Z}_{\text{I}} = Enc(\mathbf{P} \oplus \mathbf{L}), \\
    \mathbf{\hat{I}} = Dec(\mathbf{Z}_{\text{I}} \oplus \mathbf{T} \oplus \mathbf{M}).
\end{gather*}

\noindent \textbf{Flow matching~\cite{lipmanflow,liuflow}} is a framework that learns a continuous transformation between a source distribution and a target  distribution.
The source distribution is not necessarily required to be Gaussian noise, though we use Gaussian noise as a concrete example below.

During training, given a sample $X$ from the target distribution, a sampled time step $t \in [0,1]$, and a noise sample $N \sim \mathcal{N}(0, I)$ from the source distribution. An intermediate representation $X_t$ is obtained by:
\begin{gather*}
    X_t = (1 - t)\cdot X + t \cdot N.
\end{gather*}

The flow matching model is trained to estimate the velocity field $V_t$, which describes the direction from the source to the target distribution. Taking the derivative of $X_t$ with respect to $t$, we have:
\begin{gather*}
\begin{split}
     V_t &= \dfrac{dX_t}{dt} = N - X,
\end{split}
\end{gather*}
where $V_t$ indicates the direction from the source to the target distribution such that the induced flow accurately transports the source distribution to the target distribution.

Notably, while the source distribution is typically modeled as Gaussian noise in generative frameworks~\cite{esser2024scaling}, the flow matching formulation generalizes to arbitrary source distributions, provided that the source and target distributions share the same shape. In FlowTok, we directly define a unified latent space for image and text modalities, treating them as both source and target distributions. This design enables seamless generation across different modalities.
\section{Method}
\label{sec:method}

\begin{figure*}[t]
    \centering
    \includegraphics[width=1.0\linewidth]{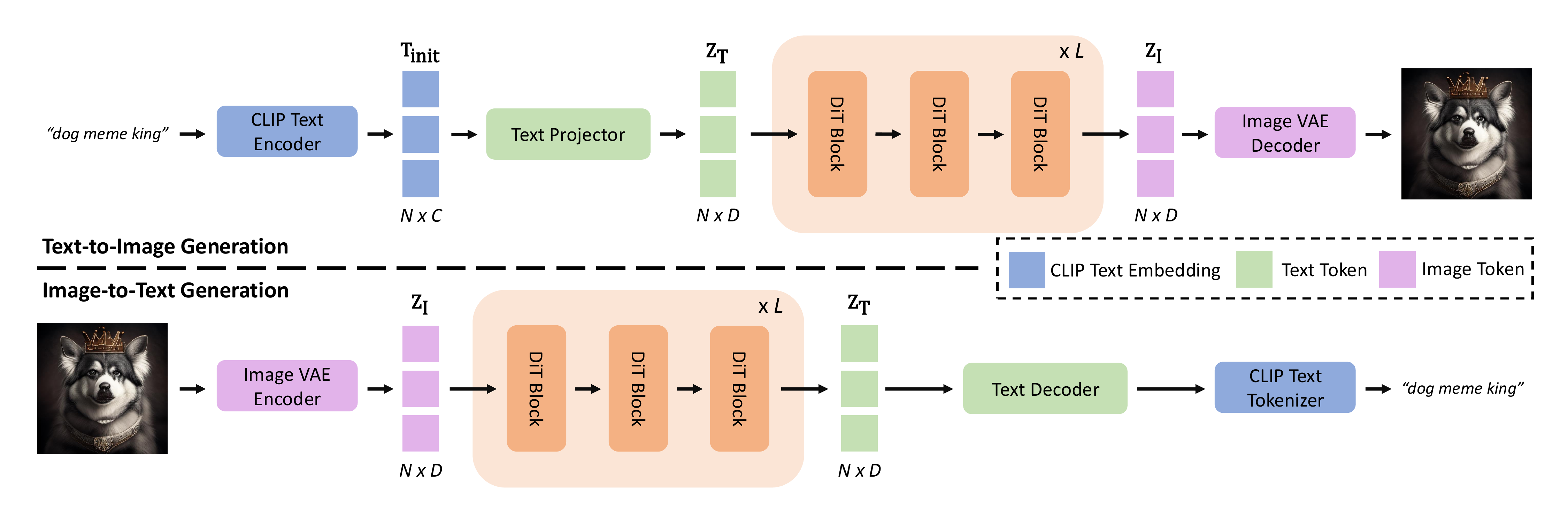}
    \caption{\textbf{Overview of \modelname.}        
    \modelname is a minimal framework that facilitates seamless flow between 1D \textcolor[HTML]{32a850}{text tokens} and \textcolor[HTML]{a8329a}{image tokens} for both text-to-image and image-to-text generation.
    \textit{Top:} For text-to-image generation, the input text is encoded by the CLIP text encoder into $\mathbf{T}_{\text{init}} \in \mathbb{R}^{N \times C}$, projected into a low-dimensional latent space as text tokens $\mathbf{Z}_{\text{T}} \in \mathbb{R}^{N \times D}$, then transformed into image tokens $\mathbf{Z}_{\text{I}} \in \mathbb{R}^{N \times D}$ of the same shape through flow matching and decoded by a 1D Image VAE Decoder to generate the final image.
    \textit{Bottom:} For image-to-text generation, an input image is encoded by a 1D Image VAE Encoder into $\mathbf{Z}_{\text{I}}$, mapped to $\mathbf{Z}_{\text{T}}$ through flow matching and decoded into text via a text decoder. Unlike conventional approaches that rely on 2D noise and image latents (\eg, $32 \times 32 \times 4$ for 256-resolution images) with text as conditions, our direct 1D transformation (\ie, $77 \times 16$) achieves a 3.3$\times$ compression rate, significantly reducing memory costs, accelerating training, and enabling faster inference.} 
   \label{fig:flowtok}
\end{figure*}


In this section, we focus on text-to-image generation as the primary task to illustrate \modelname. We first detail how images and text are projected into a unified, compact latent space as 1D tokens while preserving semantic information (\secref{subsec:unified}). Next, we introduce \modelname as a general framework for seamless flow between text and image tokens and discuss its extension to image-to-text generation under the same formulation (\secref{subsec:flowtok}).


\subsection{Unifying Latent Space of Image and Text}
\label{subsec:unified}
The structural discrepancy between text and images presents a significant challenge in unifying them within the same latent space for flow matching. Text is inherently semantic, encoded as a 1D latent sequence with high-dimensional channels to preserve meaning, whereas images contain spatially redundant information and are typically represented as 2D feature maps with lower channel dimensions to retain spatial priors.
To bridge this gap, we propose encoding images into compact 1D tokens by leveraging recent advancements in image tokenization. This formulation helps preserve the 1D structure of text embeddings, requiring only their projection into a more compressed set of tokens while ensuring that semantic information is retained. Below, we detail how both images and text are encoded.

\noindent \textbf{Encoding Images into Compact Tokens.}
We build upon the core idea of \tokenizername~\cite{kim2025democratizing} with several enhancements to improve our image tokenizer. Specifically, we replace the original learnable 1D positional embedding with RoPE~\cite{su2021roformer} to enhance \tokenizername performance. Additionally, we substitute the MLP blocks in the Vision Transformer (ViT)~\cite{dosovitskiy2020image} with SwiGLU FFN~\cite{shazeer2020glu}, which helps learn a more effective latent space~\cite{chen2024vitamin,yao2025reconstruction}. To align with the number of context tokens $N$ of the text encoder, we set the number of latent tokens $K$ in \tokenizername accordingly ($K = N = 77$ for CLIP~\cite{radford2021learning}). As a result, the encoder of \tokenizername encodes each image into a compact 1D token sequence $\mathbf{Z}_{\text{I}} \in \mathbb{R}^{K \times D}$.

\noindent \textbf{Transforming Texts into Compact Tokens.}
We use a pre-trained text encoder~\cite{radford2021learning} to extract the initial text embedding $\mathbf{T}_{\text{init}} \in \mathbb{R}^{N \times C}$, where $C$ denotes the number of channels. Notably, $C$ is typically much larger than the image latent size $D$, as it carries richer semantic information. Since our goal is to directly flow the text embedding into the image latent space, we need to ensure both embeddings have the same shape. While we already align $K$ with $N$ through careful tuning of the image tokenizer, ensuring the image is encoded to match the length of the text tokens. The remaining challenge lies in aligning the number of channels (\ie, $C$ and $D$), which we resolve using a text projector. Since only the number of channels in 
$\mathbf{T}_{\text{init}}$ needs adjustment while preserving its 1D shape, we employ a few simple Transformer blocks as the projector. To introduce variability in image generation from the same text, we model the projected text latents $\mathbf{Z}_{\text{T}} \in \mathbb{R}^{N \times D}$ as a Gaussian distribution by applying KL divergence regularization $\mathcal{L}_{\text{kld}}$. 

A crucial aspect of text-to-image generation is ensuring that the generated image accurately reflects the input text description. Since reducing the channel dimensions of text embeddings via a learnable projector may result in semantic information loss, we introduce an auxiliary text alignment loss $\mathcal{L}_{\text{align}}$ to preserve semantic consistency. 
Specifically, we employ a lightweight MLP to project $\mathbf{T}_{\text{init}}$ into a new space $\mathbf{T}_{\text{P}} \in \mathbb{R}^{N \times D}$ for alignment. We then flatten and normalize both $\mathbf{T}_{\text{P}}$ and $\mathbf{Z}_{\text{T}}$ along the channel dimension and compute a contrastive loss between them, inspired by CLIP~\cite{radford2021learning}.
Concretely, we calculate the scaled pairwise cosine similarities using a learnable temperature parameter $\tau$, followed by a symmetric cross-entropy loss:
\begin{gather*}
    \text{logits}_{\text{TZ}} = \exp(\tau) \times (\mathbf{T}_{\text{P}} \times \mathbf{Z}_{\text{T}}^\mathsf{T}),  \\
    \text{logits}_{\text{ZT}} = \exp(\tau) \times (\mathbf{Z}_{\text{T}} \times \mathbf{T}_{\text{P}}^\mathsf{T}), \\
    \mathcal{L}_{\text{align}} = (\text{CE}(\text{logits}_{\text{TZ}}, \text{labels}) + \text{CE}(\text{logits}_{\text{ZT}}, \text{labels}))/2,
\end{gather*}
where $^\mathsf{T}$ denotes the transpose operation, CE represents the cross-entropy loss, and labels are assigned based on their batch indices, ensuring that each text token is explicitly trained to align with its corresponding CLIP text embedding within the same batch. We also explore alternative approaches to preserving semantic information, such as aligning with the average-pooled text embedding or using a cosine similarity loss with a margin. However, we find that the CLIP-style loss achieves the best performance. More details are provided in~\secref{subsec:ablation}.


Through the aforementioned designs, \modelname efficiently tokenizes text into the same low-dimensional latent space while preserving semantic information. This alignment with the tokenized image latent space establishes a foundation for direct flow between compressed text tokens $\mathbf{Z}_{\text{T}}$ and image tokens $\mathbf{Z}_{\text{I}}$. Notably, when using CLIP as the text encoder, \modelname effectively reduces the latent size compared to traditional 2D flow matching methods. At an image resolution of 256, the latent size is reduced from $32 \times 32 \times 4$ to $77 \times 16$, achieving a $3.3\times$ compression. This reduction significantly lowers memory requirements and accelerates training, enhancing the framework’s efficiency and scalability.

\subsection{\modelname: A General Framework for Seamless Flow Across Text and Image Tokens}
\label{subsec:flowtok}
\noindent \textbf{Text-to-Image Generation.}
As shown in~\figref{fig:flowtok} (\textit{top}), with both image and text mapped into the same latent space, \modelname leverages vanilla flow matching~\cite{lipmanflow} by stacking DiT blocks~\cite{peebles2023scalable}. Notably, source modality (\ie, text) is directly treated as the source distribution for flow matching, removing the need for concatenation or cross-attention within the DiT blocks. This design choice further simplifies the overall framework and streamlines text-to-image generation. Combined with the compact 1D tokens introduced in~\secref{subsec:unified}, \modelname achieves high memory efficiency, supporting a batch size of 8K on 8 A100 GPUs. Additionally, it enables fast sampling, running 
over $10\times$ faster than modern text-to-image diffusion models~\cite{rombach2022high}, significantly lowering the computational barrier for training large-scale text-to-image generative models.


\noindent \textbf{Image-to-Text Generation.}
\modelname can also be seamlessly extended to image-to-text generation using compact 1D image and text tokens under the same formulation, as shown in~\figref{fig:flowtok} (\textit{bottom}). Specifically, the image tokens $\mathbf{Z}_{\text{I}}$ flow to text tokens $\mathbf{Z}_{\text{T}}$, where a trained text decoder takes $\mathbf{Z}_{\text{T}}$ as input and outputs tokenizer indices, which can then be decoded back into the corresponding caption.
\section{Experimental Results}
\label{sec:experiment}

In this section, we first provide the implementation details of \modelname (\secref{subsec:implementation}), followed by the main results on text-to-image and image-to-text generation (\secref{subsec:main}). Finally, we present the ablation studies to better understand the design choices of \modelname in text-to-image generation (\secref{subsec:ablation}). 

\subsection{Implementation Details}
\label{subsec:implementation}

\begin{table}[]
\centering
\small
\tablestyle{2.0pt}{1.0}
\scalebox{1.2}{
\begin{tabular}{l|ccccc}
model & depth & width & mlp & heads & \#params \\ \hline
\modelname-B & 12 & 768 & 3072 & 12 & 153M \\
\modelname-XL & 28 & 1152 & 4608 & 16 & 698M \\
\modelname-H & 36 & 1280 & 5120 & 20 & 1.1B 
\end{tabular}
}
\caption{
\textbf{Architecture Configuration of \modelname.}
Following prior work, we scale up DiT blocks across three configurations.
}
\label{tab:flowtok_configuration}
\end{table}

\noindent \textbf{Image Tokenizer.}  
We build our image tokenizer upon the official \tokenizername~\cite{kim2025democratizing} codebase with minimal modifications. The encoder uses ViT-B~\cite{dosovitskiy2020image}, while the decoder uses ViT-L, both operating with a patch size of $f=16$. To align with the output sequence length of CLIP’s text encoder, we set the number of 1D latent tokens $K$ to 77 and the token dimension to 16. Additionally, we enhance the tokenizer with RoPE~\cite{su2021roformer} and SwiGLU FFN~\cite{shazeer2020glu}. Notably, our enhanced tokenizer achieves a FID of 1.02 in a zero-shot evaluation on the ImageNet validation set, matching the performance of the original \tokenizername with 128 tokens.

\noindent \textbf{Text Projector.}
We train a text projector for text-to-image generation, which transforms CLIP text embeddings into a latent representation $Z_\text{T}$ of shape $77 \times 16$, aligning with the image latent space $Z_\text{I}$ encoded by our image tokenizer. The text projector consists of six Transformer~\cite{vaswani2017attention} blocks, each comprising a multi-head self-attention mechanism and a multi-layer perceptron (MLP), both enhanced with skip connections~\cite{he2016deep} to ensure stable training.

\noindent \textbf{Text Decoder.}
We train a text decoder for image-to-text generation, composed of six Transformer~\cite{vaswani2017attention} blocks, similar to the text projector. The decoder takes the text latent representation $Z_\text{T}$ as input and outputs the corresponding CLIP text tokenizer indices, which can be further converted into text using the CLIP text tokenizer.

\noindent \textbf{\modelname.}
We adopt DiT~\cite{peebles2023scalable} blocks as the fundamental building units of our \modelname to model token interactions. Specifically, we follow the DiT architecture to implement \modelname-B for efficient ablation studies and \modelname-XL for enhanced performance. To further push the performance, we scale up the depth, width, and number of attention heads, constructing \modelname-H with 1.1B parameters. The detailed model configurations are provided in~\tabref{tab:flowtok_configuration}.

\begin{table*}[!t]
\small
\centering
\tablestyle{2.0pt}{1.0}
\scalebox{1.2}{
\begin{tabular}{lc|ccc|c|c}
method & params & open-data & T$\downarrow$ & I$\uparrow$ & COCO FID-30K$\downarrow$ & MJHQ-30K FID$\downarrow$ \\ \shline
\multicolumn{7}{c}{\textit{text as conditions}} \\ \hline
GLIDE~\cite{nichol2021glide} & 5.0B & \xmark & - & - & 12.24 & - \\
Dalle$\cdot$2~\cite{ramesh2022hierarchical} & 6.5B & \xmark & - & - & 10.39 & - \\ 
LlamaGen~\cite{sun2024autoregressive} & 775M & \xmark & - & - & - & 25.59 \\
PixArt-$\alpha$~\cite{chen2024pixartalpha} & 630M & \xmark & 94.1 & 7.9 & 7.32 & 9.85 \\
SDXL~\cite{podell2023sdxl} & 2.6B & \xmark & - & - & - & 8.76 \\
LDM~\cite{rombach2022high} & 1.4B & \cmark & - & - & 12.63 & - \\
Stable-Diffusion-1.5~\cite{rombach2022high} & 860M & \cmark & 781.2 & - & 9.62 & - \\
Stable-Diffusion-2.1~\cite{rombach2022high} & 860M & \cmark & 1041.6 & - & 13.45 & 26.96 \\
Show-o~\cite{xie2024show} & 1.3B & \cmark & - & 1.0 & 9.24 & 14.99 \\
\hline
\multicolumn{7}{c}{\textit{text as source distributions}} \\ \hline
CrossFlow~\cite{liu2024flowing} & 950M & \xmark & 78.8 & 1.1 & 9.63 & - \\
\modelname-XL & 698M & \cmark & 20.4 & 22.7 & 10.06 & 7.68 \\
\modelname-H & 1.1B & \cmark & 26.1 & 18.2 & 9.67 & 7.15 \\
\end{tabular}
}
\caption{
\textbf{Zero-Shot Text-to-Image Generation Results on COCO and MJHQ-30K.}
We compare \modelname with state-of-the-art methods, categorized into two approaches: (1) \textit{text as conditions}, where text tokens are used as conditions to guide the generation process, and (2) \textit{text as source distributions}, where the model directly learns the alignment between text and image distributions.
``open-data'': Models are trained exclusively with publicly available datasets.
``T'': Model training cost, measured in 8 A100 days using float16 precision. ``I'': Model inference throughput, measured at 256px resolution in samples per second on a single A100 with batch size 64 using float16 precision.
}
\label{tab:t2i}
\end{table*}

\noindent \textbf{Dataset.}
We employ open-source datasets~\cite{kim2025democratizing} to facilitate the reproducibility of \modelname's simple framework. 
Specifically, our image tokenizer is trained on DataComp-1B~\cite{gadre2024datacomp}, and our text tokenizer is trained on COCO~\cite{lin2014microsoft}.
For text-to-image generation, inspired by recent works~\cite{dai2023emuenhancingimagegeneration,chen2024pixartalpha,chen2024pixart,zheng2024cogview3finerfastertexttoimage}, we adopt a two-stage training strategy: pre-training and fine-tuning. The pre-training stage leverages a combination of DataComp-1B~\cite{gadre2024datacomp}, CC12M~\cite{changpinyo2021conceptual}, and LAION-aesthetic~\cite{laion-aesthetic}, while the fine-tuning stage incorporates additional high-quality datasets, including LAION-art~\cite{laion-art}, LAION-pop~\cite{laion-pop}, JourneyDB~\cite{sun2024journeydb}, and DALLE3-1M~\cite{Egan_Dalle3_1_Million_2024}.
For image-to-text generation, we follow the Karpathy split~\cite{karpathy2015deep} of COCO~\cite{chen2015microsoft} to divide the training and validation sets. 
Detailed dataset information is provided in the Appendix.

\noindent \textbf{Training.} The training objectives of \modelname primarily focus on predicting velocity in flow matching, denoted as $\mathcal{L}_\text{fm}$. For text-to-image generation, we introduce two additional losses: KL-divergence loss ($\mathcal{L}_\text{kld}$) to enforce a Gaussian distribution on text tokens and text alignment loss ($\mathcal{L}_\text{align}$) to preserve semantic information as discussed in~\secref{subsec:unified}. Formally, the overall training objective is:
\begin{align*}
    \mathcal{L} = \mathcal{L}_\text{fm} + \gamma_{1} * \mathcal{L}_\text{kld} + \gamma_{2} * \mathcal{L}_\text{align},
\end{align*}
where $\gamma_1$ and $\gamma_2$ control the weighting of losses. By default, we set $\gamma_1$ to $1\times10^{-4}$ and $\gamma_2$ to $1$ for text-to-image generation, while both are set to 0 for image-to-text generation.

\noindent \textbf{Evaluation.} We follow standard evaluation practices to report relevant metrics for both text-to-image and image-to-text generation. Specifically, for text-to-image generation, we report FID-30K on the COCO~\cite{lin2014microsoft}, and FID on MJHQ-30K~\cite{li2024playground}. For image-to-text generation, we report BLEU-4~\cite{papineni2002bleu}, METEOR~\cite{banerjee2005meteor}, ROUGE~\cite{lin2004rouge}, CIDEr~\cite{vedantam2015cider}, and SPICE~\cite{anderson2016spice} on the COCO Karpathy Split~\cite{karpathy2015deep}. To incorporate classifier-free guidance (CFG)~\cite{ho2022classifier} within \modelname, we follow CrossFlow\cite{liu2024flowing} and utilize a CFG indicator. Unless otherwise stated, we find that using only 20 steps for sampling is sufficient due to the small 1D latent shape of \modelname. This significantly speeds up the inference process, enabling faster generation without compromising performance.

\begin{table}[!t]
\small
\centering
\vspace{-2mm}
\tablestyle{2.0pt}{1.0}
\scalebox{1.3}{
\begin{tabular}{l|ccccc}
method & B@4$\uparrow$ & M$\uparrow$ & R$\uparrow$ & C$\uparrow$ & S$\uparrow$ \\ 
\shline
\multicolumn{6}{c}{\textit{direct flow from image to text distributions}} \\ 
\hline
CrossFlow~\cite{liu2024flowing} & 36.4 & 27.8 & 57.1 & 116.2 & 20.4 \\ 
\modelname-XL & 37.1 & 27.8 & 57.6 & 117.0 & 20.5 \\
\hline
\multicolumn{6}{c}{\textit{other methods}}\\
\hline
MNIC~\cite{gao2019masked} & 30.9 & 27.5 & 55.6 & 108.1 & 21.0 \\
MIR~\cite{lee2018deterministic} & 32.5 & 27.2 & - & 109.5 & 20.6 \\
NAIC-CMAL~\cite{guo2020non} & 35.3 & 27.3 & 56.9 & 115.5 & 20.8 \\
SATIC~\cite{zhou2021semi} & 32.9 & 27.0 & - & 111.0 & 20.5 \\
SCD-Net~\cite{luo2023semantic} & 37.3 & 28.1 & 58.0 & 118.0 & 21.6 \\
\end{tabular}
}
\caption{
\textbf{Image-to-Text Generation Results on COCO.}
\modelname achieves performance comparable to state-of-the-art methods on image-to-text generation, evaluated on the COCO Karpathy split. For a fair comparison, we restrict our evaluation to non-autoregressive methods trained without CIDEr optimization.
}
\label{tab:i2t}
\end{table}

\begin{table*}[t!]
\centering
\subfloat[
\textbf{Text Alignment Target}
\label{tab:ablation_loss_target}
]{
\centering
\begin{minipage}{0.33\linewidth}{\begin{center}
\tablestyle{2.0pt}{1.0}
\scalebox{1.3}{
\begin{tabular}{l|c}
target & COCO FID-30K$\downarrow$ \\ \shline
Ave Pool & 36.02  \\
MLP & \final{29.14} 
\end{tabular}
}
\end{center}}\end{minipage}
}
\subfloat[
\textbf{Text Alignment Loss Function}
\label{tab:ablation_loss_function}
]{
\centering
\begin{minipage}{0.33\linewidth}{\begin{center}
\tablestyle{2.0pt}{1.0}
\scalebox{1.3}{
\begin{tabular}{l|c}
loss type & COCO FID-30K$\downarrow$ \\ \shline
Cosine & 31.80  \\
Contrastive & \final{29.14}
\end{tabular}
}
\end{center}}\end{minipage}
}
\subfloat[
\textbf{Text Alignment Loss Weight}
\label{tab:ablation_loss_weight}
]{
\centering
\begin{minipage}{0.33\linewidth}{\begin{center}
\tablestyle{2.0pt}{1.0}
\scalebox{1.3}{
\begin{tabular}{l|c}
$\gamma_2$ & COCO FID-30K$\downarrow$ \\ \shline
\final{1.0} & \final{29.14} \\
2.0 & 30.59 
\end{tabular}
}
\end{center}}\end{minipage}
}
\caption{\textbf{Ablation Studies on Text Alignment Loss.}
We conduct comprehensive ablation studies on three key aspects of the text alignment loss $\mathcal{L}_{\text{align}}$: the alignment target (\tabref{tab:ablation_loss_target}), the choice of loss function (\tabref{tab:ablation_loss_function}), and the loss weight $\gamma_2$ (\tabref{tab:ablation_loss_weight}), aiming to identify the most effective strategy for preserving semantic information within \modelname during text-to-image generation. For efficient verification, we report FID-30K on COCO using \modelname-B, without applying the CFG indicator.
}
\label{tab:ablation_loss}
\end{table*}
\subsection{Main Results}
\label{subsec:main}

\textbf{Text-to-Image Generation.} We report zero-shot text-to-image generation results on COCO~\cite{lin2014microsoft} and MJHQ-30K~\cite{li2024playground} in~\tabref{tab:t2i}. The compared methods are categorized into two groups: \textit{text as conditions}, where text serves as a guiding signal for image generation, and \textit{text as source distributions}, where text is directly modeled as a distribution in the generative process.
As observed, \modelname achieves comparable performance to prior methods in both categories on COCO FID-30K. Specifically, compared to CrossFlow~\cite{liu2024flowing}, which also uses text as the source distribution, \modelname-H attains a FID-30K of 9.67—roughly on par with CrossFlow.
When further evaluating \modelname on MJHQ-30K to assess the aesthetic quality of generated images, we find that, despite being trained solely on publicly available datasets without access to high-quality proprietary data, \modelname-XL already surpasses other state-of-the-art models, demonstrating its ability to generate diverse, high-quality images. Furthermore, \modelname-H further improves the FID score to 7.15, underscoring its superior image generation capabilities.

Beyond performance, \modelname requires significantly fewer training resources compared to existing state-of-the-art models. Specifically, \modelname-XL completes training in just 20.4 8-A100 days, while \modelname-H increases the budget slightly to 26.1 8-A100 days. In contrast, the most efficient text-as-condition model, PixArt-$\alpha$~\cite{chen2024pixartalpha}, still demands 94.1 8-A100 days. Compared to CrossFlow~\cite{liu2024flowing}, which also treats text as source distributions and requires 78.8 8-A100 days, \modelname is much more efficient.

Additionally, \modelname demonstrates significantly faster inference speeds. At 256px resolution, \modelname-XL generates 22.7 images per second, while \modelname-H achieves 18.2 images per second. In contrast, PixArt-$\alpha$ runs at 7.9 images per second, and Show-o at just 1.0 images per second. More notably, within the \textit{text as source distributions} category, \modelname achieves a 20$\times$ speedup in sampling time compared to CrossFlow, which runs at only 1.1 images per second. This efficiency stems from \modelname’s streamlined framework and its effective use of 1D tokens, significantly reducing computational overhead.

\noindent \textbf{Image-to-Text Generation.} We evaluate image-to-text generation on COCO~\cite{lin2014microsoft} using the Karpathy split~\cite{karpathy2015deep}, with results summarized in~\tabref{tab:i2t}. To ensure a fair comparison, we categorize methods into two groups: \textit{direct flow from image to text distributions}, which represents a new paradigm leveraging flow matching for direct image-to-text transformation, and \textit{other methods}, considering only those trained without CIDEr optimization.
Within the direct flow category, \modelname-XL consistently outperforms its counterpart, CrossFlow~\cite{liu2024flowing}, across most metrics. Specifically, \modelname-XL achieves a BLEU-4 (B@4) score of 37.1, surpassing CrossFlow by 0.7, and a CIDEr score of 117.0, exceeding CrossFlow by 0.8. Moreover, compared to state-of-the-art methods from other paradigms, \modelname-XL demonstrates competitive performance, highlighting direct flow matching as a promising approach for image-to-text generation. Notably, \modelname performs image-to-text generation under the same formulation using compact 1D tokens, theoretically requiring fewer training resources and enabling faster sampling compared to paradigms that operate on 2D latents, as adopted by CrossFlow. However, a direct quantitative comparison is not possible, as CrossFlow has not released the corresponding checkpoint for evaluation.

\subsection{Ablation Studies}
\label{subsec:ablation}
We conduct ablation studies on text-to-image generation using \modelname-B and evaluate them on COCO for efficiency. Our ablations focus on the design of the text alignment loss, as it plays a critical role in preserving semantic information. Specifically, we investigate three key aspects: the text alignment target (\tabref{tab:ablation_loss_target}), the choice of loss function (\tabref{tab:ablation_loss_function}), and the loss weight (\tabref{tab:ablation_loss_weight}). Details are provided below.

\noindent{\textbf{Text Alignment Target.}}
We first investigate the choice of alignment target for the projected text tokens $\mathbf{Z}_{\text{T}}$ in~\tabref{tab:ablation_loss_target}. A straightforward baseline is to directly apply average pooling (row 1) to the original CLIP text embedding $\mathbf{T}_{\text{init}}$ along the channel dimension, reducing the dimensionality from 768 to 16 to match $\mathbf{Z}_{\text{T}}$. However, this approach performs significantly worse compared to using a simple MLP to learn the alignment target (row 2), as inspired by prior works~\cite{grill2020bootstrap,oquab2023dinov2,yu2025representation}. We attribute this performance gap to the fact that adjacent channels in the CLIP text embedding are not necessarily correlated, and simple average pooling discards too much semantic information. In contrast, a learnable MLP mitigates this information loss, making it a more effective choice for defining the text alignment target.

\noindent{\textbf{Text Alignment Loss Function.}}
Next, we examine the text alignment loss function in~\tabref{tab:ablation_loss_function}. Besides the contrastive loss adopted from~\cite{radford2021learning}, we explore using a cosine similarity loss, similar to~\cite{yang2024depth}. Specifically, we compute the cosine similarity between the text tokens and the alignment target, applying a penalty to pairs with similarity below a threshold. Our experiments show that while both loss functions are effective, the contrastive loss achieves better performance. 

\noindent{\textbf{Text Alignment Loss Weight.}}
Finally, we investigate the impact of the text alignment loss weight, $\gamma_{2}$ in~\tabref{tab:ablation_loss_weight}. Our results indicate that setting $\gamma_{2}$ to 1.0, equal to the weight of the flow matching loss, is sufficient to preserve semantic information while maintaining high-quality image generation. Increasing $\gamma_{2}$ further can cause the text alignment loss to dominate the overall objective during early training stages, potentially hindering final performance. 
\vspace{4pt}
\section{Conclusion}
\vspace{4pt}
\label{sec:conclusion}
In this paper, we introduce \modelname, a minimal yet powerful framework that enables seamless direct flow between 1D text and image tokens. Through carefully designed key modules and loss functions, \modelname projects both modalities into a unified 1D latent space while preserving semantic information, enabling both text-to-image and image-to-text generation under the same formulation. This design makes \modelname highly memory-efficient, supporting an 8K batch size on just 8 A100 GPUs during training. Additionally, its simplicity accelerates convergence—within approximately 20 days on 8 A100 GPUs, \modelname achieves performance comparable to state-of-the-art models that require significantly longer training times. The streamlined design also enables over 10$\times$ faster sampling than modern text-to-image generative models. 
By releasing our code, we aim to further advance research in text-image cross-modal generation.




{
    \small
    \bibliographystyle{ieeenat_fullname}
    \bibliography{main}

@String(CVPR= {IEEE Conf. Comput. Vis. Pattern Recog.})

@String(ICCV= {Int. Conf. Comput. Vis.})

@String(ECCV= {Eur. Conf. Comput. Vis.})

@String(NIPS= {Adv. Neural Inform. Process. Syst.})

@String(ICLR = {Int. Conf. Learn. Represent.})

@String(CVPR  = {CVPR})

@String(ICCV  = {ICCV})

@String(ECCV  = {ECCV})

@String(NIPS  = {NeurIPS})

@String(ICLR  = {ICLR})

@article{ho2022classifier,
  title={Classifier-free diffusion guidance},
  author={Ho, Jonathan and Salimans, Tim},
  journal={arXiv preprint arXiv:2207.12598},
  year={2022}
}

@inproceedings{changpinyo2021conceptual,
  title={Conceptual 12m: Pushing web-scale image-text pre-training to recognize long-tail visual concepts},
  author={Changpinyo, Soravit and Sharma, Piyush and Ding, Nan and Soricut, Radu},
  booktitle={CVPR},
  year={2021}
}

@article{song2019generative,
  title={Generative modeling by estimating gradients of the data distribution},
  author={Song, Yang and Ermon, Stefano},
  journal=NIPS,
  year={2019}
}

@inproceedings{kingma2013auto,
  title={Auto-encoding variational bayes},
  author={Kingma, Diederik P and Welling, Max},
  booktitle=ICLR,
  year={2014}
}

@inproceedings{yu2025representation,
  title={Representation alignment for generation: Training diffusion transformers is easier than you think},
  author={Yu, Sihyun and Kwak, Sangkyung and Jang, Huiwon and Jeong, Jongheon and Huang, Jonathan and Shin, Jinwoo and Xie, Saining},
  booktitle={ICLR},
  year={2025}
}

@article{sun2024journeydb,
  title={Journeydb: A benchmark for generative image understanding},
  author={Sun, Keqiang and Pan, Junting and Ge, Yuying and Li, Hao and Duan, Haodong and Wu, Xiaoshi and Zhang, Renrui and Zhou, Aojun and Qin, Zipeng and Wang, Yi and others},
  journal=NIPS,
  year={2023}
}

@article{betker2023improving,
  title={Improving image generation with better captions},
  author={Betker, James and Goh, Gabriel and Jing, Li and Brooks, Tim and Wang, Jianfeng and Li, Linjie and Ouyang, Long and Zhuang, Juntang and Lee, Joyce and Guo, Yufei and others},
  journal={Computer Science},
  volume={2},
  number={3},
  pages={8},
  year={2023}
}

@misc{Egan_Dalle3_1_Million_2024,
author = {Egan, Ben and Redden, Alex and {XWAVE} and {SilentAntagonist}},
month = may,
title = {{Dalle3 1 Million+ High Quality Captions}},
howpublished = "\url{https://huggingface.co/datasets/ProGamerGov/synthetic-dataset-1m-dalle3-high-quality-captions}",
year = {2024}
}

@misc{LAION-aesthetic,
title={{LAION2B-en-aesthetic}},
howpublished = "\url{https://huggingface.co/datasets/laion/laion2B-en-aesthetic}"
}

@misc{LAION-art,
title={{LAION-art}},
howpublished = "\url{https://huggingface.co/datasets/laion/laion-art}"
}

@misc{LAION-pop,
title={{LAION-pop}},
howpublished = "\url{https://huggingface.co/datasets/laion/laion-pop}"
}

@inproceedings{lin2014microsoft,
  title={Microsoft coco: Common objects in context},
  author={Lin, Tsung-Yi and Maire, Michael and Belongie, Serge and Hays, James and Perona, Pietro and Ramanan, Deva and Doll{\'a}r, Piotr and Zitnick, C Lawrence},
  booktitle={ECCV},
  year={2014},
}

@article{chen2015microsoft,
  title={Microsoft coco captions: Data collection and evaluation server},
  author={Chen, Xinlei and Fang, Hao and Lin, Tsung-Yi and Vedantam, Ramakrishna and Gupta, Saurabh and Doll{\'a}r, Piotr and Zitnick, C Lawrence},
  journal={arXiv preprint arXiv:1504.00325},
  year={2015}
}

@article{li2024playground,
  title={Playground v2.5: Three insights towards enhancing aesthetic quality in text-to-image generation},
  author={Li, Daiqing and Kamko, Aleks and Akhgari, Ehsan and Sabet, Ali and Xu, Linmiao and Doshi, Suhail},
  journal={arXiv preprint arXiv:2402.17245},
  year={2024}
}

@inproceedings{lipmanflow,
  title={Flow Matching for Generative Modeling},
  author={Lipman, Yaron and Chen, Ricky TQ and Ben-Hamu, Heli and Nickel, Maximilian and Le, Matthew},
  booktitle={ICLR},
  year={2023}
}

@inproceedings{liuflow,
  title={Flow Straight and Fast: Learning to Generate and Transfer Data with Rectified Flow},
  author={Liu, Xingchao and Gong, Chengyue and Liu, Qiang},
  booktitle={ICLR},
  year={2023}
}

@article{ren2024flowar,
  title={FlowAR: Scale-wise Autoregressive Image Generation Meets Flow Matching},
  author={Ren, Sucheng and Yu, Qihang and He, Ju and Shen, Xiaohui and Yuille, Alan and Chen, Liang-Chieh},
  journal={arXiv preprint arXiv:2412.15205},
  year={2024}
}

@article{liu2024flowing,
  title={Flowing from Words to Pixels: A Framework for Cross-Modality Evolution},
  author={Liu, Qihao and Yin, Xi and Yuille, Alan and Brown, Andrew and Singh, Mannat},
  journal={arXiv preprint arXiv:2412.15213},
  year={2024}
}

@article{kim2025democratizing,
  title={Democratizing Text-to-Image Masked Generative Models with Compact Text-Aware One-Dimensional Tokens},
  author={Kim, Dongwon and He, Ju and Yu, Qihang and Yang, Chenglin and Shen, Xiaohui and Kwak, Suha and Chen, Liang-Chieh},
  journal={arXiv preprint arXiv:2501.07730},
  year={2025}
}

@article{polyak2024moviegencastmedia,
      title={Movie Gen: A Cast of Media Foundation Models}, 
      author={Adam Polyak and Amit Zohar and Andrew Brown and Andros Tjandra and Animesh Sinha and Ann Lee and Apoorv Vyas and Bowen Shi and Chih-Yao Ma and Ching-Yao Chuang and David Yan and Dhruv Choudhary and Dingkang Wang and Geet Sethi and Guan Pang and Haoyu Ma and Ishan Misra and Ji Hou and Jialiang Wang and Kiran Jagadeesh and Kunpeng Li and Luxin Zhang and Mannat Singh and Mary Williamson and Matt Le and Matthew Yu and Mitesh Kumar Singh and Peizhao Zhang and Peter Vajda and Quentin Duval and Rohit Girdhar and Roshan Sumbaly and Sai Saketh Rambhatla and Sam Tsai and Samaneh Azadi and Samyak Datta and Sanyuan Chen and Sean Bell and Sharadh Ramaswamy and Shelly Sheynin and Siddharth Bhattacharya and Simran Motwani and Tao Xu and Tianhe Li and Tingbo Hou and Wei-Ning Hsu and Xi Yin and Xiaoliang Dai and Yaniv Taigman and Yaqiao Luo and Yen-Cheng Liu and Yi-Chiao Wu and Yue Zhao and Yuval Kirstain and Zecheng He and Zijian He and Albert Pumarola and Ali Thabet and Artsiom Sanakoyeu and Arun Mallya and Baishan Guo and Boris Araya and Breena Kerr and Carleigh Wood and Ce Liu and Cen Peng and Dimitry Vengertsev and Edgar Schonfeld and Elliot Blanchard and Felix Juefei-Xu and Fraylie Nord and Jeff Liang and John Hoffman and Jonas Kohler and Kaolin Fire and Karthik Sivakumar and Lawrence Chen and Licheng Yu and Luya Gao and Markos Georgopoulos and Rashel Moritz and Sara K. Sampson and Shikai Li and Simone Parmeggiani and Steve Fine and Tara Fowler and Vladan Petrovic and Yuming Du},
      year={2024},
      journal={arXiv preprint arXiv:2410.13720}
}

@misc{flux2024,
    author={Black Forest Labs},
    title={FLUX},
    year={2024},
    howpublished={\url{https://github.com/black-forest-labs/flux}},
}

@article{chen2025goku,
  title={Goku: Flow Based Video Generative Foundation Models},
  author={Chen, Shoufa and Ge, Chongjian and Zhang, Yuqi and Zhang, Yida and Zhu, Fengda and Yang, Hao and Hao, Hongxiang and Wu, Hui and Lai, Zhichao and Hu, Yifei and others},
  journal={arXiv preprint arXiv:2502.04896},
  year={2025}
}

@article{sun2024autoregressive,
  title={Autoregressive Model Beats Diffusion: Llama for Scalable Image Generation},
  author={Sun, Peize and Jiang, Yi and Chen, Shoufa and Zhang, Shilong and Peng, Bingyue and Luo, Ping and Yuan, Zehuan},
  journal={arXiv preprint arXiv:2406.06525},
  year={2024}
}

@inproceedings{rombach2022high,
  title={High-resolution image synthesis with latent diffusion models},
  author={Rombach, Robin and Blattmann, Andreas and Lorenz, Dominik and Esser, Patrick and Ommer, Bj{\"o}rn},
  booktitle={CVPR},
  year={2022}
}

@misc{sdvae,
    url = {https://huggingface.co/stabilityai/sd-vae-ft-ema},
    author = {stabilityai},
    year = {2023}
}

@inproceedings{peebles2023scalable,
  title={Scalable diffusion models with transformers},
  author={Peebles, William and Xie, Saining},
  booktitle={ICCV},
  year={2023}
}

@inproceedings{chen2024pixartalpha,
title={{PixArt-$\alpha$}: Fast Training of Diffusion Transformer for Photorealistic Text-to-Image Synthesis},
author={Junsong Chen and Jincheng YU and Chongjian GE and Lewei Yao and Enze Xie and Zhongdao Wang and James Kwok and Ping Luo and Huchuan Lu and Zhenguo Li},
booktitle={ICLR},
year={2024}
}

@article{liu2024alleviating,
  title={Alleviating Distortion in Image Generation via Multi-Resolution Diffusion Models},
  author={Liu, Qihao and Zeng, Zhanpeng and He, Ju and Yu, Qihang and Shen, Xiaohui and Chen, Liang-Chieh},
  journal=NIPS,
  year={2024}
}

@article{ho2020denoising,
  title={Denoising diffusion probabilistic models},
  author={Ho, Jonathan and Jain, Ajay and Abbeel, Pieter},
  journal=NIPS,
  year={2020}
}

@inproceedings{he2016deep,
  title={Deep residual learning for image recognition},
  author={He, Kaiming and Zhang, Xiangyu and Ren, Shaoqing and Sun, Jian},
  booktitle=CVPR,
  year={2016}
}

@article{yu2024image,
  title={An Image is Worth 32 Tokens for Reconstruction and Generation},
  author={Yu, Qihang and Weber, Mark and Deng, Xueqing and Shen, Xiaohui and Cremers, Daniel and Chen, Liang-Chieh},
  journal=NIPS,
  year={2024}
}

@article{saharia2022photorealistic,
  title={Photorealistic text-to-image diffusion models with deep language understanding},
  author={Saharia, Chitwan and Chan, William and Saxena, Saurabh and Li, Lala and Whang, Jay and Denton, Emily L and Ghasemipour, Kamyar and Gontijo Lopes, Raphael and Karagol Ayan, Burcu and Salimans, Tim and others},
  journal=NIPS,
  year={2022}
}

@article{ren2025beyond,
  title={Beyond Next-Token: Next-X Prediction for Autoregressive Visual Generation},
  author={Ren, Sucheng and Yu, Qihang and He, Ju and Shen, Xiaohui and Yuille, Alan and Chen, Liang-Chieh},
  journal={arXiv preprint arXiv:2502.20388},
  year={2025}
}

@article{weber2024maskbit,
  title={MaskBit: Embedding-free Image Generation via Bit Tokens},
  author={Weber, Mark and Yu, Lijun and Yu, Qihang and Deng, Xueqing and Shen, Xiaohui and Cremers, Daniel and Chen, Liang-Chieh},
  journal={arXiv preprint arXiv:2409.16211},
  year={2024}
}

@article{podell2023sdxl,
  title={Sdxl: Improving latent diffusion models for high-resolution image synthesis},
  author={Podell, Dustin and English, Zion and Lacey, Kyle and Blattmann, Andreas and Dockhorn, Tim and M{\"u}ller, Jonas and Penna, Joe and Rombach, Robin},
  journal={arXiv preprint arXiv:2307.01952},
  year={2023}
}

@inproceedings{chang2023muse,
  title={Muse: Text-to-image generation via masked generative transformers},
  author={Chang, Huiwen and Zhang, Han and Barber, Jarred and Maschinot, AJ and Lezama, Jos{\'e} and Jiang, Lu and Yang, Ming-Hsuan and Murphy, Kevin and Freeman, William T and Rubinstein, Michael and others},
  booktitle={ICML},
  year={2023}
}

@inproceedings{gafni2022make,
  title={Make-a-scene: Scene-based text-to-image generation with human priors},
  author={Gafni, Oran and Polyak, Adam and Ashual, Oron and Sheynin, Shelly and Parikh, Devi and Taigman, Yaniv},
  booktitle={ECCV},
  year={2022}
}

@article{xie2024show,
  title={Show-o: One single transformer to unify multimodal understanding and generation},
  author={Xie, Jinheng and Mao, Weijia and Bai, Zechen and Zhang, David Junhao and Wang, Weihao and Lin, Kevin Qinghong and Gu, Yuchao and Chen, Zhijie and Yang, Zhenheng and Shou, Mike Zheng},
  journal={arXiv preprint arXiv:2408.12528},
  year={2024}
}

@inproceedings{chen2024pixart,
  title={{PixArt-$\Sigma$: Weak-to-strong training of diffusion transformer for 4k text-to-image generation}},
  author={Chen, Junsong and Ge, Chongjian and Xie, Enze and Wu, Yue and Yao, Lewei and Ren, Xiaozhe and Wang, Zhongdao and Luo, Ping and Lu, Huchuan and Li, Zhenguo},
  booktitle={ECCV},
  year={2024}
}

@article{ramesh2022hierarchical,
  title={Hierarchical text-conditional image generation with clip latents},
  author={Ramesh, Aditya and Dhariwal, Prafulla and Nichol, Alex and Chu, Casey and Chen, Mark},
  journal={arXiv preprint arXiv:2204.06125},
  volume={1},
  number={2},
  pages={3},
  year={2022}
}

@inproceedings{sohl2015deep,
  title={Deep unsupervised learning using nonequilibrium thermodynamics},
  author={Sohl-Dickstein, Jascha and Weiss, Eric and Maheswaranathan, Niru and Ganguli, Surya},
  booktitle={ICML},
  year={2015}
}

@article{yu2024randomized,
  title={Randomized Autoregressive Visual Generation},
  author={Yu, Qihang and He, Ju and Deng, Xueqing and Shen, Xiaohui and Chen, Liang-Chieh},
  journal={arXiv preprint arXiv:2411.00776},
  year={2024}
}

@article{fan2024fluid,
  title={Fluid: Scaling Autoregressive Text-to-image Generative Models with Continuous Tokens},
  author={Fan, Lijie and Li, Tianhong and Qin, Siyang and Li, Yuanzhen and Sun, Chen and Rubinstein, Michael and Sun, Deqing and He, Kaiming and Tian, Yonglong},
  journal={arXiv preprint arXiv:2410.13863},
  year={2024}
}

@inproceedings{bao2023all,
  title={All are worth words: A vit backbone for diffusion models},
  author={Bao, Fan and Nie, Shen and Xue, Kaiwen and Cao, Yue and Li, Chongxuan and Su, Hang and Zhu, Jun},
  booktitle=CVPR,
  year={2023}
}

@inproceedings{esser2024scaling,
  title={Scaling rectified flow transformers for high-resolution image synthesis},
  author={Esser, Patrick and Kulal, Sumith and Blattmann, Andreas and Entezari, Rahim and M{\"u}ller, Jonas and Saini, Harry and Levi, Yam and Lorenz, Dominik and Sauer, Axel and Boesel, Frederic and others},
  booktitle={ICML},
  year={2024}
}

@article{gadre2024datacomp,
  title={Datacomp: In search of the next generation of multimodal datasets},
  author={Gadre, Samir Yitzhak and Ilharco, Gabriel and Fang, Alex and Hayase, Jonathan and Smyrnis, Georgios and Nguyen, Thao and Marten, Ryan and Wortsman, Mitchell and Ghosh, Dhruba and Zhang, Jieyu and others},
  journal=NIPS,
  year={2023}
}

@article{yu2022scaling,
  title={Scaling autoregressive models for content-rich text-to-image generation},
  author={Yu, Jiahui and Xu, Yuanzhong and Koh, Jing Yu and Luong, Thang and Baid, Gunjan and Wang, Zirui and Vasudevan, Vijay and Ku, Alexander and Yang, Yinfei and Ayan, Burcu Karagol and others},
  journal={TMLR},
  year={2022},
}

@article{vaswani2017attention,
  title={Attention is all you need},
  author={Vaswani, Ashish and Shazeer, Noam and Parmar, Niki and Uszkoreit, Jakob and Jones, Llion and Gomez, Aidan N and Kaiser, {\L}ukasz and Polosukhin, Illia},
  journal=NIPS,
  year={2017}
}

@inproceedings{dosovitskiy2020image,
  title={An image is worth 16x16 words: Transformers for image recognition at scale},
  author={Dosovitskiy, Alexey and Beyer, Lucas and Kolesnikov, Alexander and Weissenborn, Dirk and Zhai, Xiaohua and Unterthiner, Thomas and Dehghani, Mostafa and Minderer, Matthias and Heigold, Georg and Gelly, Sylvain and Uszkoreit, Jakob and Houlsby, Neil},
  booktitle={ICLR},
  year={2021}
}

@inproceedings{radford2021learning,
  title={Learning transferable visual models from natural language supervision},
  author={Radford, Alec and Kim, Jong Wook and Hallacy, Chris and Ramesh, Aditya and Goh, Gabriel and Agarwal, Sandhini and Sastry, Girish and Askell, Amanda and Mishkin, Pamela and Clark, Jack and others},
  booktitle={ICML},
  year={2021}
}

@article{su2021roformer,
  title={RoFormer: Enhanced Transformer with Rotary Position Embedding},
  author={Su, Jianlin and Lu, Yu and Pan, Shengfeng and Murtadha, Ahmed and Wen, Bo and Liu, Yunfeng},
  journal={arXiv preprint arXiv:2104.09864},
  year={2021}
}

@article{shazeer2020glu,
  title={Glu variants improve transformer},
  author={Shazeer, Noam},
  journal={arXiv preprint arXiv:2002.05202},
  year={2020}
}

@article{dai2023emuenhancingimagegeneration, title={Emu: Enhancing Image Generation Models Using Photogenic Needles in a Haystack}, author={Xiaoliang Dai and Ji Hou and Chih-Yao Ma and Sam Tsai and Jialiang Wang and Rui Wang and Peizhao Zhang and Simon Vandenhende and Xiaofang Wang and Abhimanyu Dubey and Matthew Yu and Abhishek Kadian and Filip Radenovic and Dhruv Mahajan and Kunpeng Li and Yue Zhao and Vladan Petrovic and Mitesh Kumar Singh and Simran Motwani and Yi Wen and Yiwen Song and Roshan Sumbaly and Vignesh Ramanathan and Zijian He and Peter Vajda and Devi Parikh}, year={2023}, journal={arXiv preprint arXiv:2309.15807}}

@article{zheng2024cogview3finerfastertexttoimage,
      title={CogView3: Finer and Faster Text-to-Image Generation via Relay Diffusion}, 
      author={Wendi Zheng and Jiayan Teng and Zhuoyi Yang and Weihan Wang and Jidong Chen and Xiaotao Gu and Yuxiao Dong and Ming Ding and Jie Tang},
      year={2024},
      journal={arXiv preprint arXiv: 2403.05121},
}

@inproceedings{karpathy2015deep,
  title={Deep visual-semantic alignments for generating image descriptions},
  author={Karpathy, Andrej and Fei-Fei, Li},
  booktitle={CVPR},
  year={2015}
}

@article{albergo2022building,
  title={Building normalizing flows with stochastic interpolants},
  author={Albergo, Michael S and Vanden-Eijnden, Eric},
  journal={arXiv preprint arXiv:2209.15571},
  year={2022}
}

@inproceedings{lee2023minimizing,
  title={Minimizing trajectory curvature of ode-based generative models},
  author={Lee, Sangyun and Kim, Beomsu and Ye, Jong Chul},
  booktitle={ICML},
  year={2023}
}

@article{tong2023improving,
  title={Improving and generalizing flow-based generative models with minibatch optimal transport},
  author={Tong, Alexander and Fatras, Kilian and Malkin, Nikolay and Huguet, Guillaume and Zhang, Yanlei and Rector-Brooks, Jarrid and Wolf, Guy and Bengio, Yoshua},
  journal={arXiv preprint arXiv:2302.00482},
  year={2023}
}

@article{pooladian2023multisample,
  title={Multisample flow matching: Straightening flows with minibatch couplings},
  author={Pooladian, Aram-Alexandre and Ben-Hamu, Heli and Domingo-Enrich, Carles and Amos, Brandon and Lipman, Yaron and Chen, Ricky TQ},
  journal={arXiv preprint arXiv:2304.14772},
  year={2023}
}

@article{liu2023i2sbimagetoimageschrodingerbridge,
      title={I$^2$SB: Image-to-Image Schr\"odinger Bridge}, 
      author={Guan-Horng Liu and Arash Vahdat and De-An Huang and Evangelos A. Theodorou and Weili Nie and Anima Anandkumar},
      year={2023},
      journal={arXiv preprint arXiv:2302.05872}
}

@article{shi2023diffusion,
  title={Diffusion schr{\"o}dinger bridge matching},
  author={Shi, Yuyang and De Bortoli, Valentin and Campbell, Andrew and Doucet, Arnaud},
  journal={NeurIPS},
  year={2023}
}

@article{tong2023simulation,
  title={Simulation-free schr$\backslash$" odinger bridges via score and flow matching},
  author={Tong, Alexander and Malkin, Nikolay and Fatras, Kilian and Atanackovic, Lazar and Zhang, Yanlei and Huguet, Guillaume and Wolf, Guy and Bengio, Yoshua},
  journal={arXiv preprint arXiv:2307.03672},
  year={2023}
}

@article{zhou2023denoising,
  title={Denoising diffusion bridge models},
  author={Zhou, Linqi and Lou, Aaron and Khanna, Samar and Ermon, Stefano},
  journal={arXiv preprint arXiv:2309.16948},
  year={2023}
}

@inproceedings{chen2024vitamin,
  title={Vitamin: Designing scalable vision models in the vision-language era},
  author={Chen, Jieneng and Yu, Qihang and Shen, Xiaohui and Yuille, Alan and Chen, Liang-Chieh},
  booktitle={CVPR},
  year={2024}
}

@article{fischer2023boosting,
  title={Boosting latent diffusion with flow matching},
  author={Fischer, Johannes S and Gui, Ming and Ma, Pingchuan and Stracke, Nick and Baumann, Stefan A and Ommer, Bj{\"o}rn},
  journal={arXiv preprint arXiv:2312.07360},
  year={2023}
}

@inproceedings{bai2025meissonic,
title={Meissonic: Revitalizing Masked Generative Transformers for Efficient High-Resolution Text-to-Image Synthesis},
author={Jinbin Bai and Tian Ye and Wei Chow and Enxin Song and Qing-Guo Chen and Xiangtai Li and Zhen Dong and Lei Zhu and Shuicheng YAN},
booktitle={ICLR},
year={2025}
}

@article{gao2019masked,
  title={Masked non-autoregressive image captioning},
  author={Gao, Junlong and Meng, Xi and Wang, Shiqi and Li, Xia and Wang, Shanshe and Ma, Siwei and Gao, Wen},
  journal={arXiv preprint arXiv:1906.00717},
  year={2019}
}

@article{lee2018deterministic,
  title={Deterministic non-autoregressive neural sequence modeling by iterative refinement},
  author={Lee, Jason and Mansimov, Elman and Cho, Kyunghyun},
  journal={arXiv preprint arXiv:1802.06901},
  year={2018}
}

@article{guo2020non,
  title={Non-autoregressive image captioning with counterfactuals-critical multi-agent learning},
  author={Guo, Longteng and Liu, Jing and Zhu, Xinxin and He, Xingjian and Jiang, Jie and Lu, Hanqing},
  journal={arXiv preprint arXiv:2005.04690},
  year={2020}
}

@inproceedings{zhou2021semi,
  title={Semi-autoregressive transformer for image captioning},
  author={Zhou, Yuanen and Zhang, Yong and Hu, Zhenzhen and Wang, Meng},
  booktitle={ICCV},
  year={2021}
}

@inproceedings{luo2023semantic,
  title={Semantic-conditional diffusion networks for image captioning},
  author={Luo, Jianjie and Li, Yehao and Pan, Yingwei and Yao, Ting and Feng, Jianlin and Chao, Hongyang and Mei, Tao},
  booktitle={CVPR},
  year={2023}
}

@article{oquab2023dinov2,
  title={Dinov2: Learning robust visual features without supervision},
  author={Oquab, Maxime and Darcet, Timoth{\'e}e and Moutakanni, Th{\'e}o and Vo, Huy and Szafraniec, Marc and Khalidov, Vasil and Fernandez, Pierre and Haziza, Daniel and Massa, Francisco and El-Nouby, Alaaeldin and others},
  journal={arXiv preprint arXiv:2304.07193},
  year={2023}
}

@inproceedings{papineni2002bleu,
  title={Bleu: a method for automatic evaluation of machine translation},
  author={Papineni, Kishore and Roukos, Salim and Ward, Todd and Zhu, Wei-Jing},
  booktitle={ACL},
  year={2002}
}

@inproceedings{banerjee2005meteor,
  title={METEOR: An automatic metric for MT evaluation with improved correlation with human judgments},
  author={Banerjee, Satanjeev and Lavie, Alon},
  booktitle={Proceedings of the acl workshop on intrinsic and extrinsic evaluation measures for machine translation and/or summarization},
  year={2005}
}

@inproceedings{lin2004rouge,
  title={Rouge: A package for automatic evaluation of summaries},
  author={Lin, Chin-Yew},
  booktitle={Text summarization branches out},
  year={2004}
}

@inproceedings{vedantam2015cider,
  title={Cider: Consensus-based image description evaluation},
  author={Vedantam, Ramakrishna and Lawrence Zitnick, C and Parikh, Devi},
  booktitle={CVPR},
  year={2015}
}

@inproceedings{anderson2016spice,
  title={Spice: Semantic propositional image caption evaluation},
  author={Anderson, Peter and Fernando, Basura and Johnson, Mark and Gould, Stephen},
  booktitle=ECCV,
  year={2016}
}

@article{grill2020bootstrap,
  title={Bootstrap your own latent-a new approach to self-supervised learning},
  author={Grill, Jean-Bastien and Strub, Florian and Altch{\'e}, Florent and Tallec, Corentin and Richemond, Pierre and Buchatskaya, Elena and Doersch, Carl and Avila Pires, Bernardo and Guo, Zhaohan and Gheshlaghi Azar, Mohammad and others},
  journal=NIPS,
  year={2020}
}

@inproceedings{yang2024depth,
  title={Depth anything: Unleashing the power of large-scale unlabeled data},
  author={Yang, Lihe and Kang, Bingyi and Huang, Zilong and Xu, Xiaogang and Feng, Jiashi and Zhao, Hengshuang},
  booktitle={CVPR},
  year={2024}
}

@inproceedings{jia2021scaling,
  title={Scaling up visual and vision-language representation learning with noisy text supervision},
  author={Jia, Chao and Yang, Yinfei and Xia, Ye and Chen, Yi-Ting and Parekh, Zarana and Pham, Hieu and Le, Quoc and Sung, Yun-Hsuan and Li, Zhen and Duerig, Tom},
  booktitle={ICML},
  year={2021}
}

@article{li2019visualbert,
  title={Visualbert: A simple and performant baseline for vision and language},
  author={Li, Liunian Harold and Yatskar, Mark and Yin, Da and Hsieh, Cho-Jui and Chang, Kai-Wei},
  journal={arXiv preprint arXiv:1908.03557},
  year={2019}
}

@article{lu2019vilbert,
  title={Vilbert: Pretraining task-agnostic visiolinguistic representations for vision-and-language tasks},
  author={Lu, Jiasen and Batra, Dhruv and Parikh, Devi and Lee, Stefan},
  journal=NIPS,
  year={2019}
}

@inproceedings{cho2021unifying,
  title={Unifying vision-and-language tasks via text generation},
  author={Cho, Jaemin and Lei, Jie and Tan, Hao and Bansal, Mohit},
  booktitle={ICML},
  year={2021}
}

@article{wang2021simvlm,
  title={Simvlm: Simple visual language model pretraining with weak supervision},
  author={Wang, Zirui and Yu, Jiahui and Yu, Adams Wei and Dai, Zihang and Tsvetkov, Yulia and Cao, Yuan},
  journal={arXiv preprint arXiv:2108.10904},
  year={2021}
}

@inproceedings{li2022blip,
  title={Blip: Bootstrapping language-image pre-training for unified vision-language understanding and generation},
  author={Li, Junnan and Li, Dongxu and Xiong, Caiming and Hoi, Steven},
  booktitle={ICML},
  year={2022}
}

@inproceedings{jaegle2021perceiver,
  title={Perceiver: General perception with iterative attention},
  author={Jaegle, Andrew and Gimeno, Felix and Brock, Andy and Vinyals, Oriol and Zisserman, Andrew and Carreira, Joao},
  booktitle={ICML},
  year={2021}
}

@article{alayrac2022flamingo,
  title={Flamingo: a visual language model for few-shot learning},
  author={Alayrac, Jean-Baptiste and Donahue, Jeff and Luc, Pauline and Miech, Antoine and Barr, Iain and Hasson, Yana and Lenc, Karel and Mensch, Arthur and Millican, Katherine and Reynolds, Malcolm and others},
  journal=NIPS,
  year={2022}
}

@article{driess2023palm,
  title={Palm-e: An embodied multimodal language model},
  author={Driess, Danny and Xia, Fei and Sajjadi, Mehdi SM and Lynch, Corey and Chowdhery, Aakanksha and Wahid, Ayzaan and Tompson, Jonathan and Vuong, Quan and Yu, Tianhe and Huang, Wenlong and others},
  journal={arXiv preprint arXiv:2303.03378},
  year={2023}
}

@article{shin2025deeply,
  title={Deeply supervised flow-based generative models},
  author={Shin, Inkyu and Yang, Chenglin and Chen, Liang-Chieh},
  journal={arXiv preprint arXiv:2503.14494},
  year={2025}
}

@article{liu2025revision,
  title={Revision: High-quality, low-cost video generation with explicit 3d physics modeling for complex motion and interaction},
  author={Liu, Qihao and He, Ju and Yu, Qihang and Chen, Liang-Chieh and Yuille, Alan},
  journal={arXiv preprint arXiv:2504.21855},
  year={2025}
}

@article{ren2025grouping,
  title={Grouping First, Attending Smartly: Training-Free Acceleration for Diffusion Transformers},
  author={Ren, Sucheng and Yu, Qihang and He, Ju and Yuille, Alan and Chen, Liang-Chieh},
  journal={arXiv preprint arXiv:2505.14687},
  year={2025}
}

@article{nichol2021glide,
  title={Glide: Towards photorealistic image generation and editing with text-guided diffusion models},
  author={Nichol, Alex and Dhariwal, Prafulla and Ramesh, Aditya and Shyam, Pranav and Mishkin, Pamela and McGrew, Bob and Sutskever, Ilya and Chen, Mark},
  journal={arXiv preprint arXiv:2112.10741},
  year={2021}
}

@article{yao2025reconstruction,
  title={Reconstruction vs. Generation: Taming Optimization Dilemma in Latent Diffusion Models},
  author={Yao, Jingfeng and Wang, Xinggang},
  journal={arXiv preprint arXiv:2501.01423},
  year={2025}
}

@article{yang20241,
  title={1.58-bit FLUX},
  author={Yang, Chenglin and Liu, Celong and Deng, Xueqing and Kim, Dongwon and Mei, Xing and Shen, Xiaohui and Chen, Liang-Chieh},
  journal={arXiv preprint arXiv:2412.18653},
  year={2024}
}

@article{deng2025coconut,
  title={COCONut-PanCap: Joint Panoptic Segmentation and Grounded Captions for Fine-Grained Understanding and Generation},
  author={Deng, Xueqing and Yu, Qihang and Athar, Ali and Yang, Chenglin and Yang, Linjie and Jin, Xiaojie and Shen, Xiaohui and Chen, Liang-Chieh},
  journal={arXiv preprint arXiv:2502.02589},
  year={2025}
}

@inproceedings{ma2024sit,
  title={Sit: Exploring flow and diffusion-based generative models with scalable interpolant transformers},
  author={Ma, Nanye and Goldstein, Mark and Albergo, Michael S and Boffi, Nicholas M and Vanden-Eijnden, Eric and Xie, Saining},
  booktitle={ECCV},
  year={2024}
}

@article{
he2024a,
title={A Simple Video Segmenter by Tracking Objects Along Axial Trajectories},
author={Ju He and Qihang Yu and Inkyu Shin and Xueqing Deng and Alan Yuille and Xiaohui Shen and Liang-Chieh Chen},
journal={Transactions on Machine Learning Research},
issn={2835-8856},
year={2024},
url={https://openreview.net/forum?id=Sy6ZOStz5v},
note={}
}
}

\end{document}